\documentclass[10pt,twocolumn,letterpaper]{article}

\usepackage{cvpr}
\usepackage{times}
\usepackage{epsfig}
\usepackage{graphicx}
\usepackage{amsmath}
\usepackage{amssymb}

\usepackage{mathtools}
\usepackage{ctable}
\usepackage{hhline}
\usepackage{booktabs}
\usepackage{subcaption}
\usepackage{csquotes }

\newcolumntype{L}[1]{>{\raggedright\let\newline\\\arraybackslash\hspace{0pt}}m{#1}}
\newcolumntype{C}[1]{>{\centering\let\newline\\\arraybackslash\hspace{0pt}}m{#1}}
\newcolumntype{R}[1]{>{\raggedleft\let\newline\\\arraybackslash\hspace{0pt}}m{#1}}

\usepackage[breaklinks=true,bookmarks=false]{hyperref}

\cvprfinalcopy 

\def\cvprPaperID{3} 

\ifcvprfinal\fi
\begin{document}

\title{Multi-scale Aggregation R-CNN for 2D Multi-person Pose Estimation}

\author{Gyeongsik Moon\\
Department of ECE, ASRI\\
Seoul National University\\
{\tt\small mks0601@snu.ac.kr}
\and
Ju Yong Chang\\
Department of EI\\
Kwangwoon University\\
{\tt\small juyong.chang@gmail.com}
\and
Kyoung Mu Lee\\
Department of ECE, ASRI\\
Seoul National University\\
{\tt\small kyoungmu@snu.ac.kr}
}

\maketitle

\begin{abstract}
Multi-person pose estimation from a 2D image is challenging because it requires not only keypoint localization but also human detection. In state-of-the-art top-down methods, multi-scale information is a crucial factor for the accurate pose estimation because it contains both of local information around the keypoints and global information of the entire person. Although multi-scale information allows these methods to achieve the state-of-the-art performance, the top-down methods still require a huge amount of computation because they need to use an additional human detector to feed the cropped human image to their pose estimation model. To effectively utilize multi-scale information with the smaller computation, we propose a multi-scale aggregation R-CNN (MSA R-CNN). It consists of multi-scale RoIAlign block (MS-RoIAlign) and multi-scale keypoint head network (MS-KpsNet) which are designed to effectively utilize multi-scale information. Also, in contrast to previous top-down methods, the MSA R-CNN performs human detection and keypoint localization in a single model, which results in reduced computation. The proposed model achieved the best performance among single model-based methods and its results are comparable to those of separated model-based methods with a smaller amount of computation on the publicly available 2D multi-person keypoint localization dataset.
\end{abstract}

\section{Introduction}

Localizing semantic keypoints of an instance such as a human body or hand is an essential technique for action recognition or human-computer interaction. It has been studied for decades in computer vision community and has attracted considerable research interest.

Recently, many methods~\cite{he2017mask,chen2017cascaded,huang2017coarse,cao2016realtime,insafutdinov2016deepercut,papandreou2017towards,newell2017associative,pishchulin2016deepcut} utilize deep convolutional neural networks (CNNs) and achieved noticeable performance improvement. Although these methods have progressed considerably, they still suffer from occluded or invisible keypoints, crowded background, and high computational complexity.

\begin{figure}[t]
\begin{center}
   \includegraphics[width=1.0\linewidth]{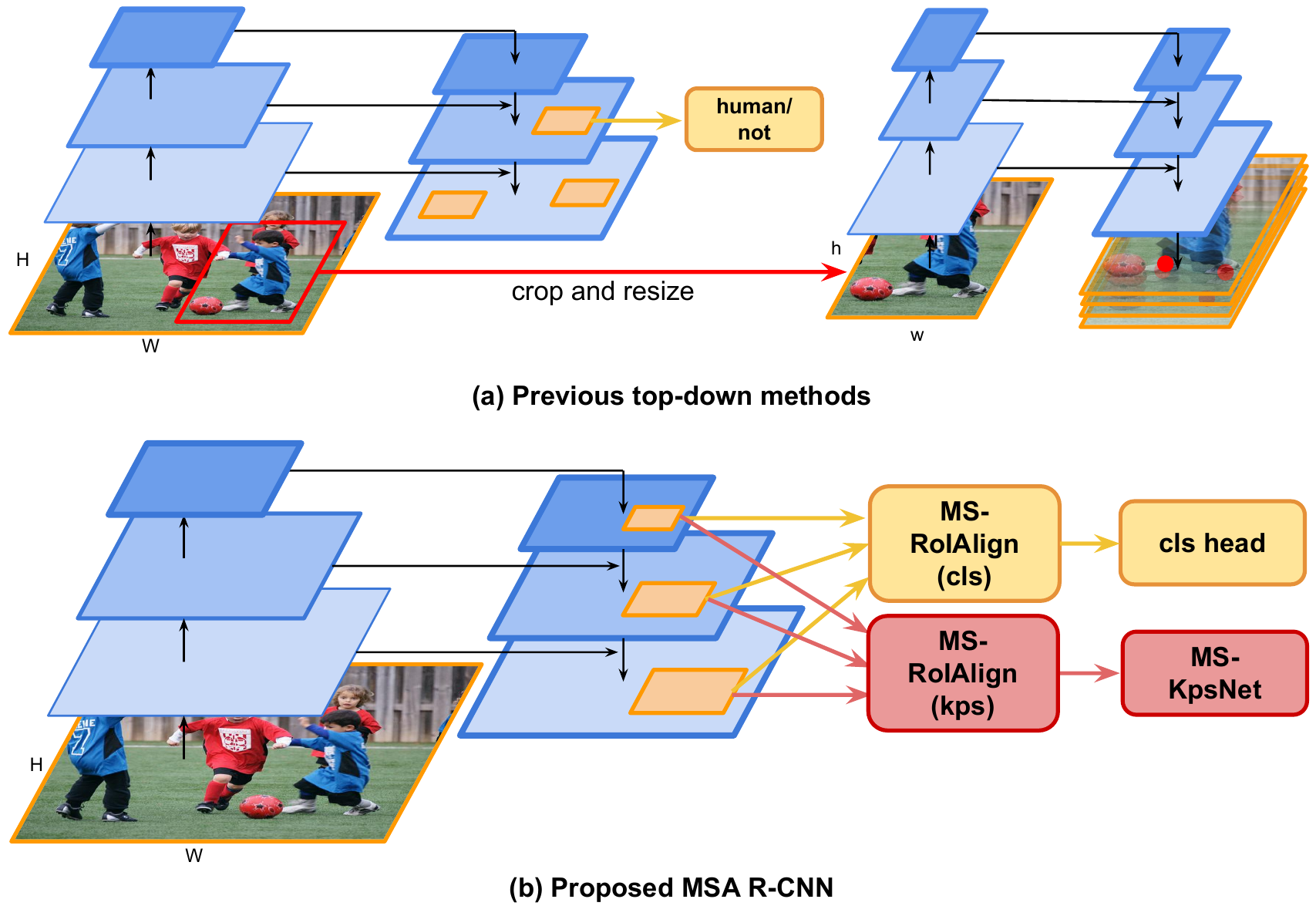}
\end{center}
   \vspace*{-5mm}
   \caption{Overall pipeline comparison with the previous top-down methods (a) and the proposed method (b). Most of the top-down approaches use two separated deep networks for multi-person pose estimation. The first model is a human detector (\textit{i.e.}, left part of (a)) and the other is a pose estimation model (\textit{i.e.}, right part of (a)). In contrast, in (b), the human detector (\textit{i.e.}, cls head) and pose estimation network (\textit{i.e.}, MS-KpsNet) are combined into a single model and share most of the feature maps.}
   \vspace*{-3mm}
\label{fig:comparison_with_top-down}
\end{figure}

In the previous top-down methods, the use of multi-scale information is crucial in performance improvement. Newell~\etal~\cite{newell2016stacked} and Chen~\etal~\cite{chen2017cascaded} used downsampling and upsampling layers with skip connections. This network architecture (\textit{i.e.}, U-net structure) is simple and effective. Huang~\etal~\cite{huang2017coarse} aggregated multi-scale information by concatenating feature maps from multiple scale spaces. Although these multi-scale approaches exhibit state-of-the-art accuracy, they require a huge amount of computation because they need to use an additional human detector to feed the cropped human image to their model. Considering that both of the recent state-of-the-art object detectors~\cite{ren2015faster,he2017mask} and keypoint localization networks~\cite{chen2017cascaded,huang2017coarse,papandreou2017towards} are primarily based on the very deep backbone networks~\cite{he2016deep,xie2017aggregated}, the total amount of computation is very large. 

\begin{figure}[t]
\begin{center}
   \includegraphics[width=1.0\linewidth]{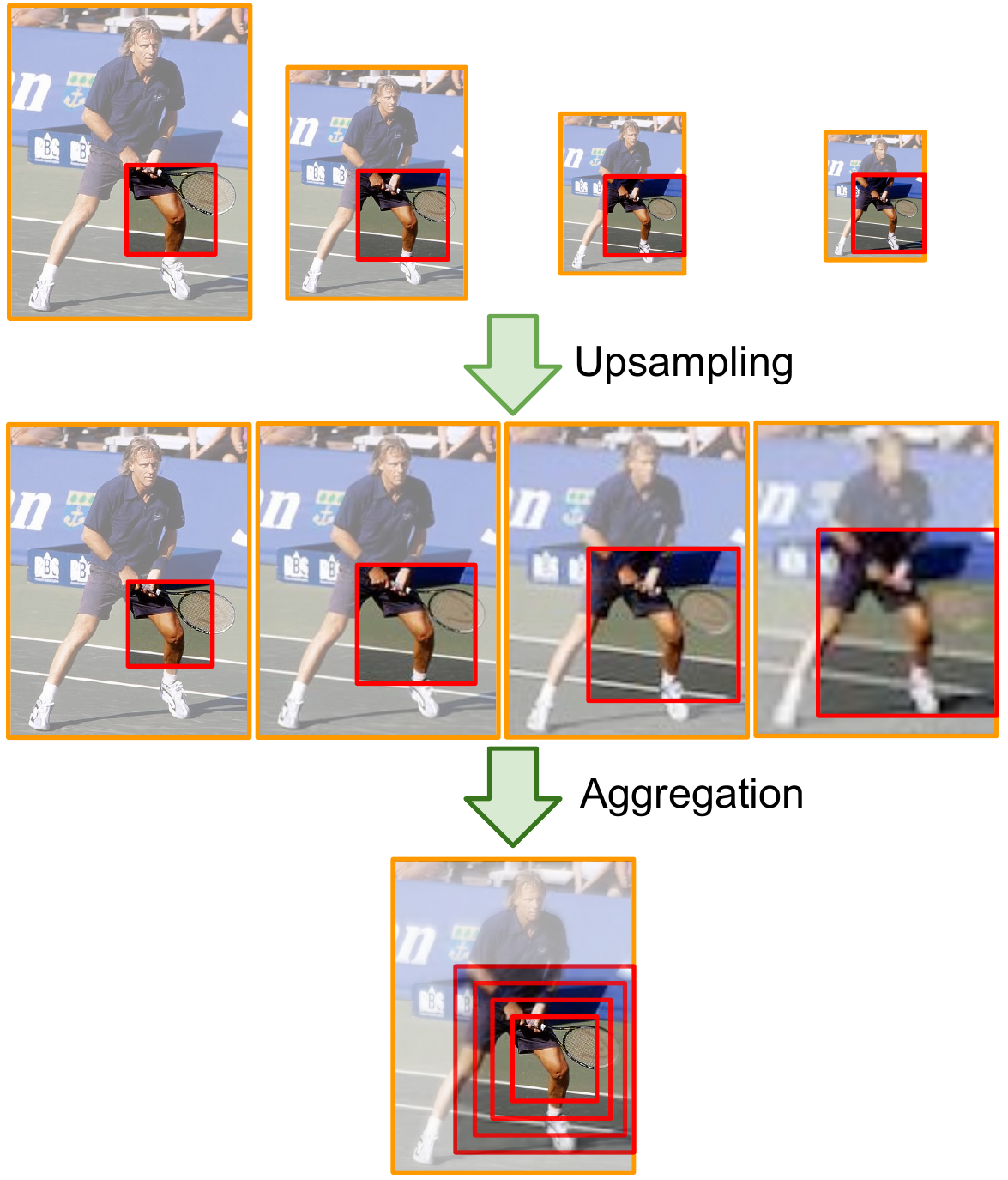}
\end{center}
   \vspace*{-5mm}
   \caption{The MSA R-CNN extracts multi-scale information from downsampled and upsampled feature maps and aggregates the information by using MS-RoIAlign and MS-KpsNet. The orange and red boxes denote the extracted feature maps of the human and the receptive fields of convolutional layers, respectively. We take an example of the left knee area.}
   \vspace*{-3mm}
\label{fig:multi-scale}
\end{figure}

By contrast, Mask R-CNN~\cite{he2017mask} learns human detection and keypoint localization in a single model that can be trained in an end-to-end manner. Based on the shared feature maps, two small separated head networks for human/non-human classification and keypoint localization are jointly learned to minimize the weighted sum of loss functions. However, this method does not fully utilize multi-scale information which is a bottleneck to the accurate keypoint localization. Specifically, RoIAlign~\cite{he2017mask} extracts a feature of each proposal from a single-scale feature map by considering the size of each proposal. The small proposals are extracted from a fine-scaled feature map, while the large proposals from a coarse-scaled feature map. However, because \textit{each} proposal is from a \textit{single-scale} feature map, RoIAlign fails to fully exploit multi-scale information. Also, the keypoint head network consists of several sequentially added convolutional layers. As this design gradually increases receptive field size, the output feature would mainly focus on global information rather than local information. This makes it hard to aggregate multi-scale information.

To remedy the heavy computation in the previous top-down methods~\cite{chen2017cascaded,huang2017coarse,pishchulin2016deepcut} and the lack of multi-scale information in the Mask R-CNN~\cite{he2017mask}, we propose a multi-scale aggregation R-CNN (MSA R-CNN). The MSA R-CNN crops and resizes human bounding box proposals from feature maps instead of an input image as shown in Figure~\ref{fig:comparison_with_top-down}. This property enables the MSA R-CNN to share feature maps for human detection and keypoint localization, which results in considerably reduced computation. Also, to exploit multi-scale information more effectively, we propose multi-scale RoIAlign block (MS-RoIAlign) and multi-scale keypoint head network (MS-KpsNet). In contrast to the original RoIAlign, the MS-RoIAlign obtains human proposals from multi-scale feature maps instead of a single feature map and aggregates them. It enables the model to exploit various scales of the feature maps which is helpful for the final prediction. Also, the MS-KpsNet obtains human proposals from the MS-RoIAlign and estimates heatmaps for each keypoint by utilizing multi-scale information. The proposed MS-KpsNet consists of downsampling and upsampling layers with residual skip connections which help incorporate local- and global-scale information. To summarize, both of the MS-RoIAlign and MS-KpsNet try to extract and aggregate multi-scale information as in Figure~\ref{fig:multi-scale}.

We validated the usefulness of the MS-RoIAlign and MS-KpsNet on the MS COCO keypoint detection dataset~\cite{lin2014microsoft}. The experimental results show that the proposed items (\textit{i.e.}, MS-RoIAlign and MS-KpsNet) bring large performance improvement. Our model outperforms all single model-based methods and achieves comparable results to those of separated model-based methods but with less computation on a challenging benchmark~\cite{lin2014microsoft}.

Our contributions can be summarized as follows:

\begin{itemize}
\item The MSA R-CNN reduces a large amount of computation compared with other top-down methods by combining human detection and keypoint localization in a single model.
\item The MS-RoIAlign and MS-KpsNet effectively utilize multi-scale information, thereby enhancing performance.
\item Our model achieved the best performance among single model-based methods and comparable results to those of separated model-based methods on the MS COCO keypoint detection dataset~\cite{lin2014microsoft}.
\end{itemize}

\section{Related works}
The proposed method is closely related to the following two tracks. In this paper, we mainly focus on methods based on the CNN.

\begin{figure*}
\begin{center}
\includegraphics[width=1.0\linewidth]{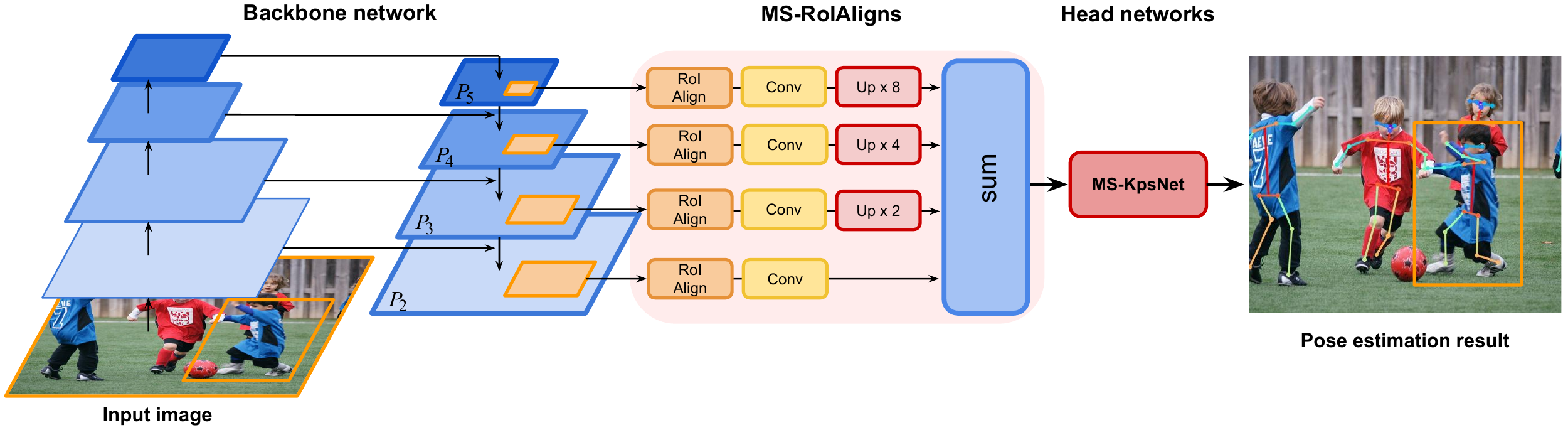}
\end{center}
   \vspace*{-3mm}
   \caption{Overall pipeline of the proposed method. The input image that contains multiple humans is fed to the backbone network. After the backbone network generates human bounding box proposals, the features of the proposals are extracted by the MS-RoIAlign from the multiple feature maps. The extracted features are aggregated and passed to the head networks in a parallel manner. We exclude the MS-RoIAlign and head network of the classification from the figure, and only one proposal with orange rectangle is drawn for simplicity.}
   \vspace*{-3mm}
\label{fig:overall_pipeline}
\end{figure*}

\textbf{Single-person pose estimation.}
Toshev~\etal~\cite{toshev2014deeppose} directly estimated the Cartesian coordinates of body joints by using a multi-stage deep network and obtained remarkable performance. Tompson~\etal~\cite{tompson2014joint} estimated the per-pixel likelihood for each joint by using CNN and used it as the unary term in an external graphical model to accurately estimate joint positions. Liu~\etal~\cite{wei2016convolutional} utilized multiple stages of refinement to enlarge receptive fields. Newell~\etal~\cite{newell2016stacked} proposed a stacked U-Net structure model (\textit{i.e.}, hourglass structure) to exploit information from multiple scales. Bulat and Tzimiropoulos~\cite{bulat2016human} adopted a detection subnetwork to help the regression subnetwork accurately localize body joints. Carreria~\etal~\cite{carreira2016human} proposed an iterative error feedback-based human pose estimation system. It is supervised to progressively refine the initial pose to the groundtruth pose. Chu~\etal~\cite{chu2017multi} enhanced the stacked hourglass network~\cite{newell2016stacked} by incorporating multi-context attention mechanism. Chen~\etal~\cite{chen2017adversarial} also improved the hourglass network~\cite{newell2016stacked} with the adversarial loss to generate plausible poses.

\textbf{Multi-person pose estimation.}
There are two streams in the multi-person pose estimation. The first one, top-down approach, relies on a human detector which predicts bounding boxes of humans. The detected human is cropped and fed to the pose estimation network. The second one, bottom-up approach, localizes all the human body keypoints in an input image and groups them using proposed clustering algorithms of each work. 

~\cite{he2017mask,chen2017cascaded,huang2017coarse,papandreou2017towards,xiao2018simple,Moon_2019_CVPR_PoseFix} are based on the top-down approach. Papandreou~\etal~\cite{papandreou2017towards} estimated heatmaps and offsets for each joint. The offsets are defined as vectors toward the groundtruth joint location from each tensor grid. He~\etal~\cite{he2017mask} proposed Mask R-CNN which can perform human detection and keypoint localization in a single model. It extracts human features from a feature map instead of an input image by using RoIAlign. Chen~\etal~\cite{chen2017cascaded} used a coarse-to-fine approach and designed a network called cascaded pyramid network (CPN) which consists of GlobalNet and RefineNet. The GlobalNet is U-Net shaped model and supervised to estimate heatmaps for each keypoint from each scale of a feature map. The RefineNet is designed to refine the localization output from the GlobalNet by focusing on hard keypoints. Xiao~\etal~\cite{xiao2018simple} proposed a straightforward architecture-based human pose estimation model.

~\cite{cao2016realtime,insafutdinov2016deepercut,newell2017associative,pishchulin2016deepcut,kocabas2018multiposenet} are based on the bottom-up approach. DeepCut~\cite{pishchulin2016deepcut} assigned the detected keypoints to different persons in an image by formulating the assignment problem as an integer linear program. DeeperCut~\cite{pishchulin2016deepcut} improves the DeepCut~\cite{pishchulin2016deepcut} by introducing image-conditioned pairwise terms. Cao~\etal~\cite{cao2016realtime} proposed part affinity fields (PAFs) that models the relationship between human body keypoints and assembled the localized keypoints using the estimated PAFs. Newell~\etal~\cite{newell2017associative} introduced a pixel-wise tag value to assign localized keypoints to a certain human. Kocabas~\etal~\cite{kocabas2018multiposenet} proposed a pose residual network for assigning detected keypoints to each person.

\section{Overview of the proposed model}

The proposed MSA R-CNN has three components. The first is a single backbone network for shared feature extraction. The second component is separated into two MS-RoIAligns for human/non-human classification and keypoint localization. The outputs of MS-RoIAligns are fed to two small head networks (\textit{i.e.}, classification head network and MS-KpsNet) which are the third component of our system. The backbone network extracts deep features, and each MS-RoIAlign passes these features to the corresponding head network. The classification head network predicts whether a proposal is human or not, and the MS-KpsNet estimates heatmaps for each joint. The overall pipeline is visualized in Figure~\ref{fig:overall_pipeline}.

\section{Backbone network for shared features}

The feature pyramid network (FPN)~\cite{lin2017feature} is adopted as the backbone network. The FPN extracts deep features using ResNet~\cite{he2016deep} or ResNeXt~\cite{xie2017aggregated} and gradually upsamples the features. Each upsampled feature is summed by lateral connections with the feature map in the same scale space from the front part of the network. This upsampling with skip connection architecture is widely used for dense prediction such as segmentation~\cite{ronneberger2015u} and keypoint localization~\cite{newell2016stacked} because it can provide more semantic information to fine-scale feature maps. Following ~\cite{ren2015faster}, the backbone network is supervised to generate human bounding box proposals from an input image by using a binary cross entropy loss for each sampled feature map grid and a smooth L1 loss to refine the bounding box coordinates.

\begin{figure}[t]
\begin{center}
   \includegraphics[width=1.0\linewidth]{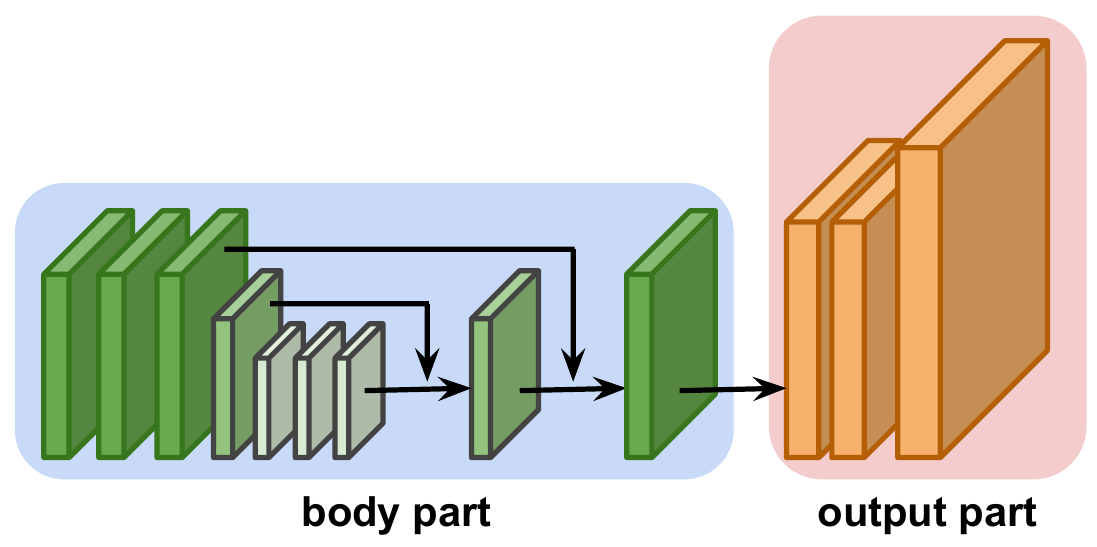}
\end{center}
   \vspace*{-3mm}
   \caption{Architecture of the MS-KpsNet. It consists of a body and output parts. The feature map in the body part passes through convolutional, downsampling and upsampling layers. In the output part, the feature map from the body part is upsampled by a deconvolutional layer and bilinear interpolation is applied for accurate estimation. The loss is calculated on the four times upsampled RoI which is the last feature map of the output part.}
   \vspace*{-3mm}
\label{fig:MS-KpsNet}
\end{figure}

\section{Multi-scale RoIAlign block (MS-RoIAlign)}

The MS-RoIAlign passes the extracted human feature from the backbone network to the corresponding head network. 

The original RoIAlign~\cite{he2017mask} extracts human proposal features from a single feature map. The feature map is selected among several scales according to the size of the proposal~\cite{lin2017feature}. The original method assigns small and large proposals to large feature maps (fine-scale, low-level feature maps) and small feature maps (coarse-scale, high-level feature maps), respectively. However, this straightforward assignment strategy can result in sub-optimal performance. For example, two proposals that have almost the same area can be assigned to two different feature maps. Such an assignment can make learning unstable because the proposals have similar areas. Hence, we consider feature maps from the entire-scale space instead of a single-scale feature map. Another disadvantage of the original RoIAlign is that other levels of feature maps are discarded. Exploiting multi-level feature maps provides more information than exploiting only a single feature map. The low-level features contain detailed local information, which results in high localization accuracy in the fine-scale space. Furthermore, the high-level features have rich semantic information resulting from the large receptive field size in the coarse-scale space. Compared with the existing RoI assignment strategy~\cite{lin2017feature}, the proposed MS-RoIAlign can utilize all information from multi-level feature maps.

\begin{table*}
\small
\centering
\setlength\tabcolsep{1.0pt}
\def\arraystretch{1.1}
\begin{tabular}{L{3.7cm}C{1cm}C{1cm}C{1cm}C{1cm}C{1cm}|C{1.3cm}C{1.3cm}C{1.3cm}C{1.3cm}C{1.3cm}C{1.3cm}}
\specialrule{.1em}{.05em}{.05em}
    Methods   & $AP^{kps}$ & $AP^{kps}_{.50}$ & $AP^{kps}_{.75}$ & $AP^{kps}_M$ & $AP^{kps}_L$ &  $AP^{bb (H)}$ & $AP^{bb (H)}_{.50}$ & $AP^{bb (H)}_{.75}$ & $AP^{bb (H)}_S$ & $AP^{bb (H)}_M$ & $AP^{bb (H)}_L$\\ \hline
Baseline      & 64.1 & 86.4 & 69.3 & 58.9 & 72.7 & 52.7 & 82.3 & 57.4 & 35.6 & 60.5 & 68.7 \\ 
 + Only from $P_2$      & 64.4 & 86.4 & 69.9 & 59.4 & 72.8 & 52.5 & 82.2 & 57.2 & 35.4 & 60.7 & 68.2 \\ 
 + 1$\times$1 conv output      & 64.7 & 86.3 & 70.4 & 59.6 & 73.2 & 52.4 & 82.5 & 56.9 & 35.2 & 60.6 & 68.0 \\ 
 + MS-KpsNet     & 66.2 & 87.0 & 72.7 & 61.3 & 74.5 & 52.6 & 82.3 & 57.4 & 35.6 & 60.6 & 68.3 \\
 + Longer training    & 66.5 & 87.5 & 72.5 & 61.5 & 75.0 & 53.4 & 82.8 & 58.3 & 36.1 & 61.5 & 69.3 \\ 
 + MS-RoIAlign   & 67.4 & 87.7 & 73.5 & 62.1 & 76.0  & 54.8 & 83.4 & 60.1 & 37.4 & 62.8 & 71.2 \\ 
 + Average of Top-2s   & 67.6 & 87.7 & 73.7 & 62.5 & 76.1  & 54.8 & 83.4 & 60.1 & 37.4 & 62.8 & 71.2 \\ 
 + Test-time augmentation   & \textbf{70.3} & \textbf{89.2} & \textbf{76.6} & \textbf{65.9} & \textbf{77.9}  & \textbf{56.4} & \textbf{84.9} & \textbf{61.8} & \textbf{39.0} & \textbf{64.2} & \textbf{72.6} \\ \specialrule{.1em}{.05em}{.05em}
    & \textbf{+6.2} & \textbf{+2.8} & \textbf{+7.3} & \textbf{+7.0} & \textbf{+5.2} & \textbf{+3.7} & \textbf{+2.6} & \textbf{+4.4} & \textbf{+3.4} & \textbf{+3.7} & \textbf{+3.9} \\
\end{tabular}
\vspace*{-3mm}
\caption{Effect of various settings in terms of the performance on the MS COCO validation set. $AP^{bb (H)}$ means the average precision of detection task for the human class only.}
\vspace*{-3mm}
\label{table:ablation_total}
\end{table*}

The pipeline of the MS-RoIAlign is visualized in Figure~\ref{fig:overall_pipeline}. The MS-RoIAlign extracts ($2^{n+3}$$\times$$2^{n+3}$, $2^{n+2}$$\times$$2^{n+2}$, $2^{n+1}$$\times$$2^{n+1}$, $2^{n}$$\times$$2^{n}$)-sized RoIs from upsampled feature maps ($P_2$, $P_3$, $P_4$, $P_5$) for each proposal. The extracted features go through convolutional layers followed by subsequent upsampling layers. The RoIs are resized to a fixed size (\textit{i.e.}, $2^{n+3}$$\times$$2^{n+3}$) and aggregated by summation. Then, it is fed to the corresponding head network. This procedure lets the following head networks fully utilize the multi-scale features instead of narrowing the choice to a single-scale feature. The $n$ is set to 0 for the classification and 1 for the keypoint localization to make the RoI sizes similar to those of the Mask R-CNN~\cite{he2017mask}. Except for the parameter $n$ related to the size of the input RoI, the MS-RoIAligns of the two tasks have exactly the same architecture. The small difference in the RoI sizes of our method and Mask R-CNN~\cite{he2017mask} makes no difference in terms of the performance.

\begin{table}
\small
\centering
\setlength\tabcolsep{1.0pt}
\def\arraystretch{1.1}
\begin{tabular}{L{2.0cm}C{1.2cm}C{2.5cm}C{1.8cm}}
\specialrule{.1em}{.05em}{.05em} 
   Aggregation  & $AP^{kps}$ & Num of params & Train mem \\ \hline
Sum  & 67.6 & 76.3M & 10.8 GB  \\ 
Concat  &  67.6 & 137.2M & 14.4 GB   \\ \specialrule{.1em}{.05em}{.05em} 
\end{tabular}
\vspace*{-3mm}
\caption{Performance comparison of the MS-RoIAlign with different aggregation method. The AP is from the test result of the MS COCO validation set. The train mem indicates the required amount of GPU memory in the training stage.}
\vspace*{-3mm}
\label{table:MS-RoIAlign}
\end{table}

\section{Multi-scale keypoint head network (MS-KpsNet)}

The human proposal features extracted by the MS-RoIAlign are fed to the proposed MS-KpsNet which predicts heatmaps for each keypoint. To effectively utilize both of the local- and global-scale information, the MS-KpsNet is designed with downsampling and upsampling architectures and residual skip connections.

The architecture of the MS-KpsNet is presented in Figure~\ref{fig:MS-KpsNet}. The MS-KpsNet starts from three consecutive convolutional layers and goes through two rounds of downsampling. Each downsampling layer is followed by a convolutional layer. The downsampled feature passes two convolutional layers and subsequently upsampled followed by a residual skip connection. The forward is finished after two rounds of upsampling and skip connection. Like the downsampling layers, a convolutional layer is added after each residual skip connection in the upsampling part. Max pooling with stride and kernel size of 2 is used for downsampling layers and nearest neighbor with a scale factor of 2 is used for upsampling layers. The skip connection is a single convolutional layer. All the convolutional layers have 3$\times$3 kernels and are followed by the activation function (\textit{i.e.}, ReLU).  Cross-entropy loss function $L$ is calculated after softmax normalization as follows:
\begin{equation}
L = -\frac{1}{N}\sum_{n=1}^{N}\sum_{i,j}H_n^*(i,j)\log H_n(i,j),
\end{equation}
where $H_{n}^{*}$ and $H_{n}$ are the groundtruth and estimated heatmaps with softmax applied for $n$th keypoint, respectively, and $N$ denotes the number of keypoints. Groundtruth heatmap $H_{n}^{*}$ is encoded as a one-hot representation.

\section{Implementation details}
Our model is based on the official Caffe2~\cite{jia2014caffe} implementation of the Mask R-CNN~\cite{Detectron2018}. Following the Mask R-CNN~\cite{he2017mask}, human bounding box proposals are generated from an independently trained RPN ~\cite{lin2017feature,ren2015faster} for convenient ablation study and fair comparison. Note that it can be trained in an end-to-end manner and achieves slightly better results compared with the model trained from independently trained RPN~\cite{Detectron2018}. 

\textbf{Training.}
Our model is based on ResNet-50~\cite{he2016deep} and all weights are initialized with a publicly released model pre-trained on the ImageNet dataset~\cite{russakovsky2015imagenet}. We adopt image-centric training~\cite{girshick2015fast}. For each image, we sample 512 RoIs with positive-to-negative ratio of 1:3. For data augmentation, the length of the short side of an image is randomly sampled between 640 and 800 pixels. Weight decay and momentum are set to 0.0001 and 0.9, respectively. As we used two GPUs that are smaller than that of the Mask R-CNN~\cite{he2017mask}, we used the linear scaling rule~\cite{goyal2017accurate} to set the learning rate and number of iterations according to the number of GPUs. Each GPU takes 2 images to generate a mini-batch. For the classification head network, we used the same loss function (i.e., binary cross-entropy) and architecture (i.e., two fully-connected layers) as the Mask R-CNN~\cite{he2017mask}.

\begin{table*}
\small
\centering
\setlength\tabcolsep{1.0pt}
\def\arraystretch{1.1}
\begin{tabular}{L{3.1cm}C{2.5cm}C{1.1cm}C{1.1cm}C{1.1cm}C{1.1cm}C{1.1cm}C{1.1cm}C{1.1cm}C{1.1cm}C{1.1cm}C{1.1cm}}
\specialrule{.1em}{.05em}{.05em}
      Methods  & Backbone & $AP^{kps}$ & $AP^{kps}_{.50}$ & $AP^{kps}_{.75}$ & $AP^{kps}_M$ & $AP^{kps}_L$ & $AR^{kps}$ & $AR^{kps}_{.50}$ & $AR^{kps}_{.75}$ & $AR^{kps}_M$ & $AR^{kps}_L$\\ \hline
\multicolumn{12}{l}{\textbf{\textit{Separated model-based methods}}}  \\
RMPE~\cite{fang2017rmpe}    & - &61.0 & 82.9 & 68.8 & 57.9 & 66.5 & - & - & - & - & - \\  

G-RMI~\cite{papandreou2017towards}    & ResNet-101 & 64.9 & 85.5 & 71.3 & 62.3 & 70.0 & 69.7 & 88.7 & 75.5 & 64.4 & 77.1 \\
CPN~\cite{chen2017cascaded}   & ResNet-Inception & 72.1 & \textbf{91.4} & \textbf{80.0} & 68.7 & \textbf{77.2} & \textbf{78.5} & \textbf{95.1} & \textbf{85.3} & \textbf{74.2} & \textbf{84.3} \\ 
CFN~\cite{huang2017coarse}    & Inception v2 & \textbf{72.6} & 86.1 & 69.7 & \textbf{78.3} & 64.1 & - & - & - & - & - \\ \hline
\multicolumn{12}{l}{\textbf{\textit{Single model-based methods}}}  \\
CMU-Pose~\cite{cao2016realtime}   & -  & 61.8 & 84.9 & 67.5 & 57.1 & 68.2 & 66.5 & 87.2 & 71.8 & 60.6 & 74.6 \\ 
Mask R-CNN~\cite{he2017mask}   & ResNet-50-FPN  & 63.1 & 87.3 & 68.7 & 57.8 & 71.4 & - & - & - & - & - \\ 
AE~\cite{newell2017associative}     & - & 65.5 & 86.8 & 72.3 & 60.6 & 72.6 & 70.2 & 89.5 & 76.0 & 64.6 & 78.1 \\ 
\textbf{MSA R-CNN (Ours)}    & ResNet-50-FPN & \textbf{68.2} & \textbf{89.7} & \textbf{75.0} & \textbf{63.8} & \textbf{75.6} & \textbf{74.4} & \textbf{93.4} & \textbf{80.3} & \textbf{69.2} & \textbf{81.5} \\  \specialrule{.1em}{.05em}{.05em}
\end{tabular}
\vspace*{-3mm}
\caption{Comparison with the state-of-the-art methods on the MS COCO test-dev set. Methods that involve extra training data or use ensemble technique are excluded.}
\vspace*{-3mm}
\label{table:comparison_with_sota}
\end{table*}

\begin{table}
\small
\centering
\setlength\tabcolsep{1.0pt}
\def\arraystretch{1.1}
\begin{tabular}{L{3.1cm}C{1.0cm}C{2.0cm}C{2.0cm}}
\specialrule{.1em}{.05em}{.05em} 
   Method &  $AP_{kps}$ & Running time & Train mem \\ \hline
CPN-50 & 67.3 & 0.10 + 0.21 & 7.9 + 10.5 GB   \\
Mask R-CNN & 63.1 & 0.17 & 7.7 GB   \\
Mask R-CNN+ & 67.0 & 0.31 & 14.2 GB   \\
\textbf{MSA R-CNN (Ours)} & 67.6 & 0.21 & 10.8 GB   \\ \specialrule{.1em}{.05em}{.05em}
\end{tabular}
\vspace*{-3mm}
\caption{Computational complexity comparison with the state-of-the-art top-down methods. The AP is from the test result on the MS COCO validation set. The running time is the number of seconds required to process an image and the train mem indicates the amount of the GPU memory consumption in the training stage. For CPN-50, the former and latter results are from the human detector and the pose estimation model, respectively.}
\vspace*{-3mm}
\label{table:computational_complexity}
\end{table}

\textbf{Inference.}
At test time, the extracted RoI bounding boxes pass the classification head network and the estimated bounding box refinement vector refines the coordinates of the bounding boxes. Then, the refined bounding boxes pass the MS-KpsNet, which differs from the parallel computation used in training. This sequential prediction speeds up inference and improves accuracy due to the use of fewer and more accurate RoIs. The predicted heatmaps for each body keypoint are resized to the original RoIs and the position of the highest response for each keypoint is identified and warped to the final result of our model.

All the hyper-parameters are adopted from Mask R-CNN~\cite{he2017mask} and FPN~\cite{lin2017feature} without any fine-tuning.
\section{Experiment}
\subsection{Dataset and evaluation metric}
The proposed model is trained on the MS COCO~\cite{lin2014microsoft} training set which includes 57K images and 150K person instances. The validation is performed on the MS COCO validation set which includes 5K images and testing is conducted on the test-dev set that includes 20K images. Following the public benchmark, we used the object keypoint similarity (OKS)~\cite{ronchi2017benchmarking} based mAP as an evaluation metric. The OKS defines the similarity between the coordinates of two human body keypoints which is similar to intersection over union in object detection.

\subsection{Ablation study}
We trained our model on the MS COCO training set and validated the proposed components on the MS COCO validation set.

\textbf{Multi-scale aggregation network.}
To demonstrate the validity of the multi-scale aggregation, we compared the performances of the baseline model~\cite{he2017mask} and the proposed MSA R-CNN in Table~\ref{table:ablation_total}. 

\textbf{1) Baseline model.} We employed ResNet-50-based Mask R-CNN~\cite{he2017mask} as the baseline model.  

\textbf{2) Only from $P_2$.} When all human bounding box proposals are extracted from the finest-scale feature map ($P_2$), not from the assigned feature map according to size~\cite{lin2017feature}, the performance is slightly improved. This may be because the keypoint localization task prefers features from large upsampled feature maps. Although the extracted RoI size is the same regardless of whether it is from the $P_2$ to the $P_5$, the detailed local and fine-scaled information from the $P_2$ is helpful for accurate keypoint localization. 

\textbf{3) 1$\times$1 conv output.} In the output part of the original keypoint head network, we used a deconv layer followed by an 1$\times$1 conv layer to generate heatmaps for each joint instead of a single deconv layer. This is to separate the two tasks, which upsample the feature map and estimate the heatmap.

\textbf{4) MS-KpsNet.} When the proposed MS-KpsNet is introduced, the performance increases by 1.5 AP. This shows the usefulness of the MS-KpsNet that is designed to effectively utilize multi-scale information. 

\textbf{5) Longer training.} For stable convergence, we scaled the training schedule by approximately 1.44 times, which slightly improves performance. 

\textbf{6) MS-RoIAlign.} The MS-RoIAlign increases performance by 0.9 AP which shows utilizing multi-scale features is better than relying on a single-scale feature. 

\textbf{7) Average of Top-2s.} To improve performance of the keypoint localization at high precision thresholds, we select top-2 grid with the highest probability from the estimated heatmap. Then, the weighted average of the locations of the selected grids based on their probability becomes the final location of each keypoint. 

\textbf{8) Test-time augmentation.} The multi-scale test-time augmentation is commonly used to boost the performance~\cite{newell2017associative,cao2016realtime}. It averages heatmaps from multiple sizes of an input image, which makes the model robust to scale variations. 

All the proposed methods obtain 6.2 AP improvement compared with the baseline model.

\textbf{Aggregation method.}
We explore the best aggregation method in the MS-RoIAlign in terms of the performance and computational complexity. We compared our aggregation method (\textit{i.e.}, summation) with concatenation which is used in CFN~\cite{huang2017coarse}. When concatenation is used, the upsampled RoIs are concatenated along the channel dimension. The first three feature map dimension of the MS-KpsNet are changed to 1024, 1024, and 512 in response to the increased number of channels. As Table~\ref{table:MS-RoIAlign} shows, there is a marginal performance difference between concatenation and summation although concatenation requires more parameters and consumes more GPU memory in the training stage. Therefore, we used summation as the aggregation method in the MS-RoIAlign.

\subsection{Comparison with the state-of-the-art methods}
We compared the performance of the MSA R-CNN on the MS COCO ~\cite{lin2014microsoft} test-dev set with that of recent state-of-the-art methods including RMPE~\cite{fang2017rmpe}, CMU-Pose~\cite{cao2016realtime}, Mask R-CNN~\cite{he2017mask}, Associative Embedding (AE)~\cite{newell2017associative}, CFN~\cite{huang2017coarse}, G-RMI~\cite{papandreou2017towards}, and CPN~\cite{chen2017cascaded}. Table~\ref{table:comparison_with_sota} shows the performance comparison. 

Our MSA R-CNN outperforms all the single model-based methods. We additionally tried to compare the proposed MSA R-CNN with a recently introduced single model-based method, MultiPoseNet~\cite{kocabas2018multiposenet}. As they only reported the performance using ensemble on the test-dev set, we compare our MSA R-CNN with the MultiPoseNet~\cite{kocabas2018multiposenet} on the validation set without ensemble and testing time augmentation. Our ResNet-50-based model achieves 67.6 mAP while their ResNet-50-based model achieves 62.3 mAP. Moreover, their model with deeper backbone network (\textit{i.e.}, ResNet-101) achieves 63.9 mAP which is still lower than ours. This comparison clearly shows the proposed MSA R-CNN outperforms all the single model-based methods.

On the other hand, the proposed method performs slightly worse than recent state-of-the-art top-down methods~\cite{huang2017coarse,chen2017cascaded} that require an additional human detection model. As our model contains both of the human detector and keypoint localization network, a limit exists in the use of computational resources such as GPU memory, which poses a limitation in obtaining better performance. This prevents us from utilizing well-known factors for performance boosting such as a deeper backbone network~\cite{lin2017feature}. 

By contrast, separated model-based methods train and test the human detector and pose estimation model separately. Accordingly, a computational resource limitation exists for each model and not the combined model. The increased computational resource limitation can be used for performance enhancement. For example, recent state-of-the-art top-down methods use very deep network-based human detectors~\cite{he2017mask,ren2015faster,peng2018megdet}, which consume a large amount of computation resource. The CPN used human detection results from the MegDet~\cite{peng2018megdet} which is trained on 128 GPUs. The MegDet~\cite{peng2018megdet} obtains 50.5 AP on the MS COCO~\cite{lin2014microsoft} detection validation set for all classes whereas our baseline (\textit{i.e.}, ResNet-50-based Mask R-CNN) obtains 37.3 AP. Moreover, their keypoint localization models not only can use very deep backbone networks including ResNeXt~\cite{xie2017aggregated} and ResNet-Inception~\cite{szegedy2017inception}, but also can be designed with as highly sophisticated network architecture~\cite{huang2017coarse,chen2017cascaded}. 

\begin{figure*}
\begin{center}
\includegraphics[width=0.95\linewidth]{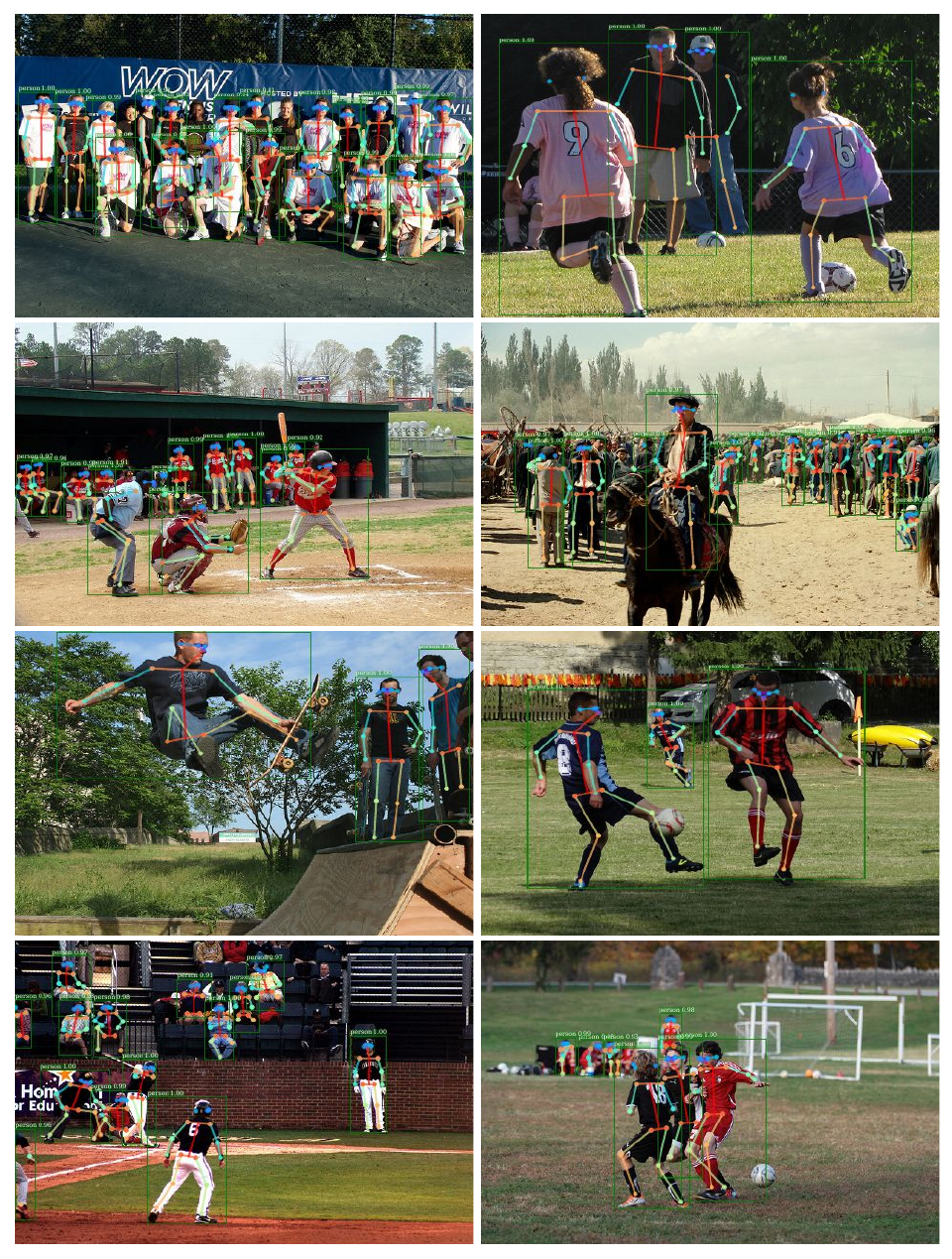}
\end{center}
   \vspace*{-8mm}
   \caption{Qualitative results of our MSA R-CNN on the MS COCO test-dev dataset.}
   \vspace*{-3mm}
\label{fig:qualitative_result}
\end{figure*}

Figure~\ref{fig:qualitative_result} shows the qualitative results of our MSA R-CNN on the MS COCO~\cite{lin2014microsoft} keypoint detection test-dev set.

\subsection{Computational complexity}
We compared the accuracy and computational complexity of the proposed method with those of the Mask R-CNN, very deep backbone based-Mask R-CNN~\cite{he2017mask} (\textit{i.e.}, Mask R-CNN+) and the basic model of the CPN~\cite{chen2017cascaded} (\textit{i.e.}, CPN-50) in Table~\ref{table:computational_complexity}. Among the separated model-based methods, we chose the CPN because it released the code and achieved top performance. The Mask R-CNN+ uses ResNeXt-101-FPN~\cite{xie2017aggregated,lin2017feature} as a backbone network and CPN-50 is based on the ResNet-50~\cite{he2016deep}. We use the same backbone based object detector with ours (\textit{i.e.}, ResNet-50-FPN-based Mask R-CNN) as the human detector of the CPN-50 because the human detector code of the CPN (i.e., MegDet~\cite{peng2018megdet}) is unavailable. For a fair comparison, ensembling, and keypoint rescoring, and test time augmentation techniques are excluded.

As Table~\ref{table:computational_complexity} shows, our method achieves the best accuracy with the least amount of computational resource in both of the training and testing stages compared with the Mask R-CNN+ and CPN-50. The CPN-50 requires 48\% longer running time in the testing stage and 70\% more GPU memory in the training stage to achieve similar accuracy with the MSA R-CNN. Considering that the CPN-50 is the simplest model of the CPN with a basic human detector, previous separated model-based systems require a huge amount of computation to achieve the state-of-the-art performance. 

Furthermore, compared with the Mask R-CNN, the MSA R-CNN increases the computational complexity by approximately 30\% whereas the Mask R-CNN+ increases it by around 80\% in both of the training and testing stages. This result indicates that the proposed modules in the MSA R-CNN (\textit{i.e.}, MS-RoIAlign and MS-KpsNet) efficiently increases accuracy compared with using deeper backbone network which is the most widely used strategy for accuracy improvement~\cite{he2017mask,he2016deep,lin2017feature}.

\section{Conclusion}

We proposed a novel and powerful network, MSA R-CNN, for 2D multi-person pose estimation. In contrast to previous top-down methods, the proposed method performs human detection and keypoint localization in a single model. This unified model allows us to save a large amount of computations compared with the separated model-based methods. Also, to effectively utilize multi-scale information, the MS-RoIAlign and MS-KpsNet are proposed, which extract multi-scale features and aggregate them. MS-RoIAlign and MS-KpsNet obtain remarkable performance improvements. Our method outperforms all the existing single model-based methods and achieved comparable results to those of the separated model-based methods on the challenging benchmark. Codes will be released for reproduction.

\section*{Acknowledgments}
This work was supported by deep learning-based human pose estimation system development project from NAVER.

\clearpage

{\small
\bibliographystyle{ieee_fullname}
\bibliography{egpaper_final}
}

\end{document}